\pdfoutput=1

\documentclass{article}

\usepackage{arxiv}

\usepackage[utf8]{inputenc} 
\usepackage{authblk}

\usepackage[T1]{fontenc}    
\usepackage{url}            
\usepackage{booktabs}       
\usepackage{amsfonts}       
\usepackage{nicefrac}       
\usepackage{microtype}      
\usepackage{lipsum}

\usepackage{subcaption}
\usepackage{array}
\usepackage{amsmath}
\usepackage{graphicx}

\title{A Spiking Neural Network \\for Image Segmentation}

\author[1]{Kinjal Patel}
\author[1]{Eric Hunsberger}
\author[2]{Sean Batir \thanks{This work was done when the author was at BMW Technology Office, 2606 Bayshore Parkway, Mountain View CA 94043}}
\author[1,3]{Chris Eliasmith}
\affil[1]{Applied Brain Research, Inc}
\affil[2]{Army Research Lab}
\affil[3]{Centre for Theoretical Neuroscience, University of Waterloo}
\affil[1]{{\{{kinjal.patel, eric.hunsberger, chris.eliasmith}\}@appliedbrainresearch.com}}
\affil[2]{sbatir@alum.mit.edu}

\begin{document}
\maketitle
\vspace{1cm}
\begin{abstract}
We seek to investigate the scalability of neuromorphic computing for computer vision, with the objective of replicating non-neuromorphic performance on computer vision tasks while reducing power consumption. We convert the deep Artificial Neural Network (ANN) architecture U-Net to a Spiking Neural Network (SNN) architecture using the Nengo framework. Both rate-based and spike-based models are trained and optimized for benchmarking performance and power, using a modified version of the ISBI 2D EM Segmentation dataset consisting of microscope images of cells. We propose a partitioning method to optimize inter-chip communication to improve speed and energy efficiency when deploying multi-chip networks on the Loihi neuromorphic chip. We explore the advantages of regularizing firing rates of Loihi neurons for converting ANN to SNN with minimum accuracy loss and optimized energy consumption. We propose a percentile based regularization loss function to limit the spiking rate of the neuron between a desired range. The SNN is converted directly from the corresponding ANN, and demonstrates similar semantic segmentation as the ANN using the same number of neurons and weights. However, the neuromorphic implementation on the Intel Loihi neuromorphic chip is over 2x more energy-efficient than conventional hardware (CPU, GPU) when running online (one image at a time). These power improvements are achieved without sacrificing the task performance accuracy of the network, and when all weights (Loihi, CPU, and GPU networks) are quantized to 8 bits.
\end{abstract}


\section{Introduction}

Researchers generally acknowledge that the biological brain possesses remarkable energy efficiency, especially given the number of computations that it is estimated to perform daily.  This comes from comparing the computational efficiency of the brain to that of man-made computers. For example, the expected metabolic profile of an adult human brain over a period of 24 hours is 420 kcal, which converts to 488 daily watt hours, or an average of 20.4 watts \cite{metabolic}. In contrast, modern GPU hardware uses on the order of hundreds of watts, and is often fully utilized running full-scale video processing in real-time, which still only represents a fraction of the computation performed by the human brain. 

Biological inspiration has been a driving factor in the development of many modern computer vision and other signal processing techniques. For example, convolutional neural networks (CNNs) were inspired by both the hierarchical processing and the relatively translation-invariant local processing of the brain's visual system \cite{cnn}. However, a key limitation is the energy cost required to achieve state-of-the-art object recognition and segmentation accuracy, associated with running these models on traditional (CPU/GPU) hardware. These observations regarding efficiency and functionality support the claim that the real-time information processing capabilities of biological neural systems still significantly exceed contemporary implementations of hardware-based neural networks. Core to the discipline of neuromorphic engineering is the idea that biological systems can inspire hardware design choices. However, it remains an ongoing challenge to demonstrate that many such hardware designs can scale to reasonably sized problems, while exhibiting both competitive task performance as well as reduced energy use.

One core hypothesis of several neuromorphic designs is that the heavy computational cost of contemporary deep artificial neural networks (ANNs) results from the continual transmission of real-valued activities between connected nodes in the network, as well as the subsequent matrix multiplication or convolution. Consequently, implementing neural networks with neural spiking may enable the same information transmission and function, but decrease the costs of signal transmission and computation. Binary-valued spikes both reduce the number of bits per transmission by turning real-valued signals into binary ones, and they make signals sparse in time by not transmitting information for each connection every timestep. They also allow the multiply-and-accumulate typically inherent in matrix multiplication or convolution to be turned into simply accumulation (since spikes are zero or one, a spike times any value $x$ is either zero or $x$, avoiding the need for explicit multiplication). Therefore, mimicking spiking---the energy-efficient signaling mechanism used by most of the biological human brain---may lead to more energy-efficient artificial neural network implementations. 

One example of spike-based neuromorphic hardware is the Intel Loihi neuromorphic chip \cite{davies2018loihi}. This novel design is purposely built to run asynchronous spiking networks, while allowing on-line, customizable learning at very low power. The chip includes 128 cores, each supporting up to 1024 neurons, fabricated with Intel's 14 nm process. The mesh protocol of the chip allows up to 16,384 chips to be used in a single system. Conveniently, it is supported by the Nengo neural network development tool \cite{bekolay2014} to allow for efficient, multi-chip programmability.

The energy efficiency of Loihi has been tested through numerous different networks over the past few years. Here, we will focus on feed-forward architectures, since they are most applicable to the image-processing problems that we are looking to address. For some applications, Loihi shows a clear advantage in dynamic power usage when compared with conventional hardware like CPU and GPU, though the time required to process each input is often higher on Loihi. Examples of feed-forward networks demonstrating lower power on Loihi include: 
\begin{itemize}
    \item A keyword spotting network \cite{blouw2018keyword}, which occupies 8 cores (one 16$^\text{th}$ of a Loihi chip) and achieved as low as 109$\times$ less energy than a GPU and 23$\times$ less energy than a CPU;
    \item A unidimensional SLAM network \cite{slam-loihis} with input from distance and speed encoders, which occupies 82 cores (about two-thirds of a Loihi chip) and achieves similar accuracy to the widely used GMapping algorithm with 100$\times$ less energy compared to a CPU;
    \item A container classification and slip detection network \cite{visual-tactile} using an event-based camera and NeuTouch fingertip sensor, which occupies about 19 cores (about one-seventh of a Loihi chip) and achieves 162$\times$ energy reduction compared with a GPU;    
    \item A gesture recognition network \cite{hand-gesture} combining event based camera input and electromyography (EMG) signals, achieving a 30$\times$ reduction in energy compared to GPU for the full network;
    \item An image retrieval network \cite{image-retrieval} for the FashionMNIST dataset, which occupies 17 cores (about one-eighth of a Loihi chip) and uses between 1.3$\times$ \emph{more} power and 4.3$\times$ less power on Loihi as compared to a variety of different CPUs and GPUs.
\end{itemize}
None of these networks use more than one Loihi chip. Furthermore, the networks performing image-processing tasks mostly use input from event-based cameras, which is already temporally sparse. Only the image retrieval network uses standard images as input, and it shows the thinnest margins for energy efficiency of Loihi over standard hardware.

In this paper, we investigate both the scalability of deep neural networks on Loihi and the applicability of Loihi to processing standard images, by implementing an image segmentation network spanning multiple Loihi chips. The simple and straightforward U-Net architecture \cite{unet} makes it well suited for this objective. We leverage the Nengo framework to translate a simplified U-Net into a spiking network to deploy on the Intel Loihi neuromorphic chip. The simplified U-Net consists of 238K neurons and occupies two chips utilizing a total of 237 cores, making it one of the largest examples of a functional neuromorphic application implemented on Loihi. We optimize architecturally identical rate- and spike-based models to process a modified version of the ISBI 2D EM Segmentation dataset \cite{isbi}, consisting of HeLa cell images with minor obfuscation. We devise a custom loss function to regularize the firing rates of the neurons and analyse the effects of regularising firing rates on accuracy and energy efficiency. We also propose a partitioning algorithm to deploy SNN on Loihi which minimizes inter-chip communication resulting in faster and energy efficient network. We benchmark both the energy efficiency and the task performance accuracy of our implementation on neuromorphic hardware (Loihi) and conventional hardware using both online (one image at a time) and parallel (batch) processing. 
\section{Method}
\label{sec:method}
The following subsections describe the details of the original and downsized U-Net architecture, followed by the data augmentation, network implementation, and power benchmarking methods used. 
\\
\subsection{Network Architecture}
\label{subsec:nw_architecture}
\begin{figure}[h!]
\begin{center}
    \includegraphics[width=15cm,trim={1cm 0cm 0cm 0cm},clip,keepaspectratio]{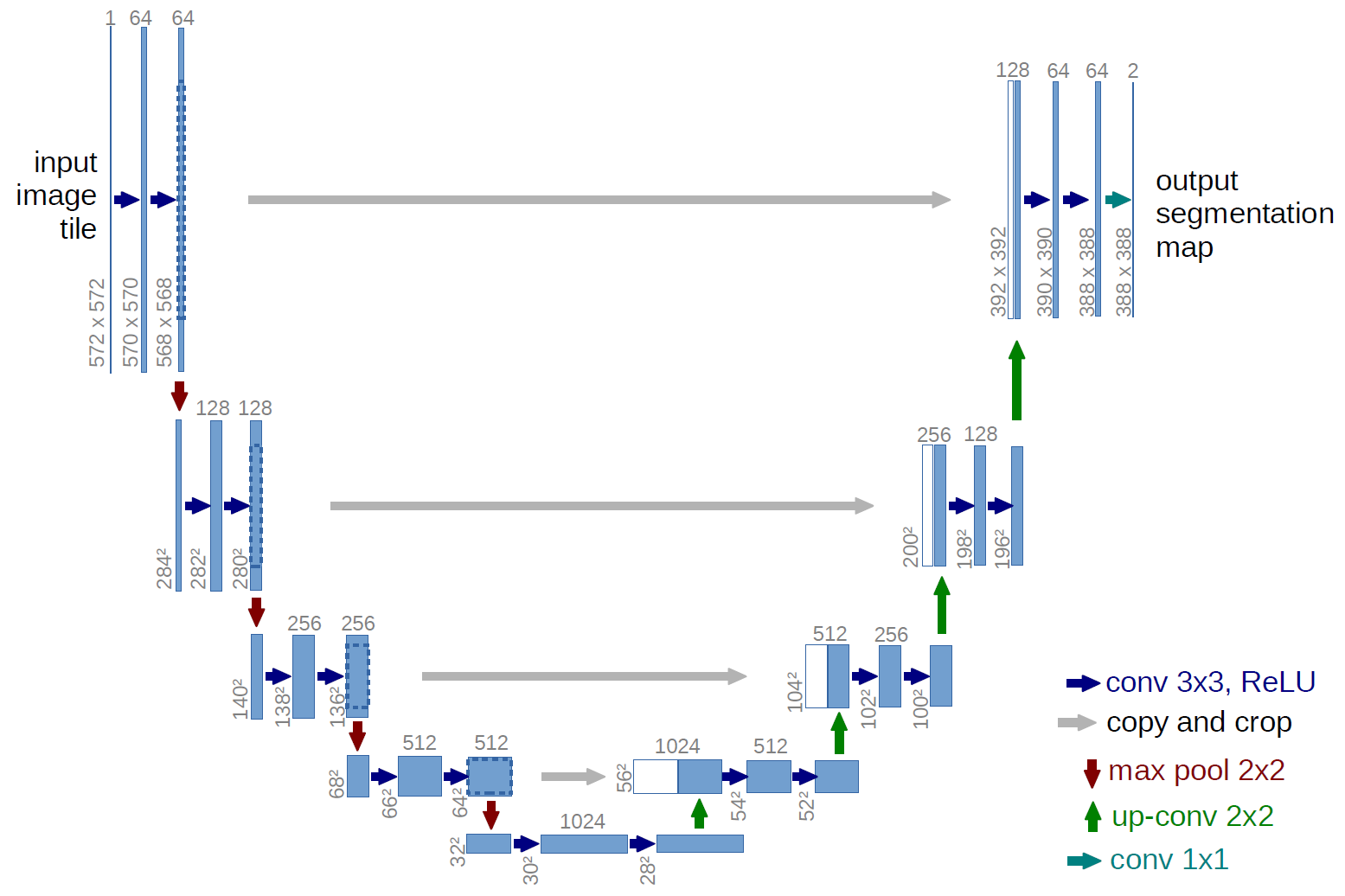}
\end{center}
    \caption{Original U-Net architecture for semantic segmentation \cite{unet}}
    \label{fig:unet_original}
\end{figure}

The original U-Net architecture is illustrated in Figure \ref{fig:unet_original}. It is comprised of a contractive part (left side) and an expansive part (right side). The contractive part can be broken up into subunits, where each subunit begins with either the input image or a downsampled version of the previous subunit output (downsampling uses max-pooling with $2 \times 2$ strides), followed by two convolution layers (non-padded) each with a $3 \times 3$ kernel and a ReLU non-linearity. The expansive part is also made up of subunits, each doing approximately the inverse of a contractive subunit: two convolution layers (non-padded) each with a $3 \times 3$ kernel and a ReLU non-linearity, followed by a transposed convolution (i.e.,  a deconvolution) layer with $2 \times 2$ strides. The output of this subunit is then concatenated with the cropped output from the corresponding contractive subunit, to provide input to the next expansive subunit. These “copy and crop” connections are a key innovation of the U-Net architecture, as they allow it to process information at multiple scales, and recombine it in a meaningful way. In our model, we describe all layers operating at the same scale as one “meta-layer,” which consists of “pre” (contractive) and “post” (expansive) groups of layers. 

\begin{figure}[h!]
\begin{center}
    \includegraphics[width=10cm,keepaspectratio]{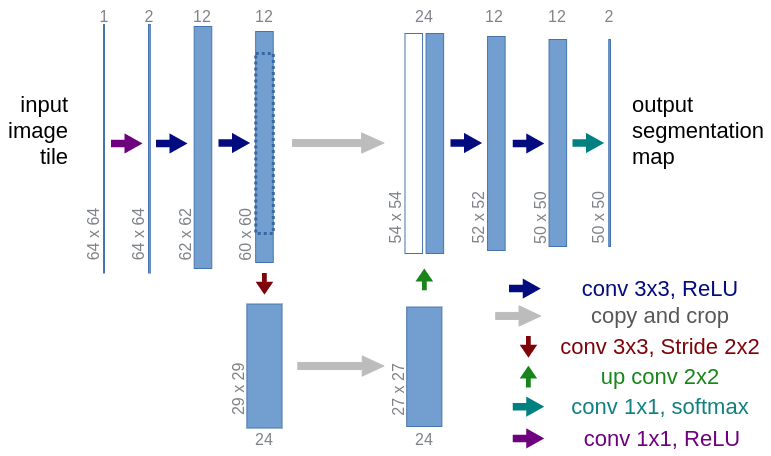}
\end{center}
    \caption{Scaled-down U-Net architecture with reduced input size, number of layers, number of filter channels with a total of 238K neurons}
    \label{fig:unet_scaled-down}
\end{figure}

We aim to assess the scalability of the network and energy consumption on an embedded platform like KapohoBay comprising of 262K neurons over two Loihi chips. To meet this size constraint, we scaled-down the original U-Net architecture comprising of 137M neurons. The scaled-down U-Net architecture is detailed in Figure \ref{fig:unet_scaled-down}. We reduce the size of the input image from $572 \times 572$ to $64 \times 64$, and reduce the number of meta-layers from five to two. We also reduce the number of feature channels in the first meta-layer from 64 to 12, resulting in 24 channels in the second meta-layer. We also add an input encoder layer which converts input image into spikes using a trainable 1x1 convolution layer. This network ended up using 238K neurons in the final spiking and non-spiking implementations. At this scale, it is one of the larger functional networks to run on spiking hardware, and to our knowledge the largest vision system on Loihi. 

As with many CNNs, the U-Net includes max pooling layers. However, there is no general technique for effectively implementing max pooling in spiking networks. Consequently, we replace the max pooling connection with a convolutional connection with a stride of 2 ($3 \times 3$ filters, ReLU nonlinearity, no padding). Critically, the simplified network contains all of the computations required of the full-scale U-Net (e.g., convolution, concatenation, convolution transpose, and copy \& crop) except max pooling.

\\
\subsection{Data Augmentation}
\label{subsec:data_augmentation}

\begin{figure}[h]
\centering
\begin{subfigure}{.3\textwidth}
    \centering
    \includegraphics[width=0.9\linewidth,keepaspectratio]{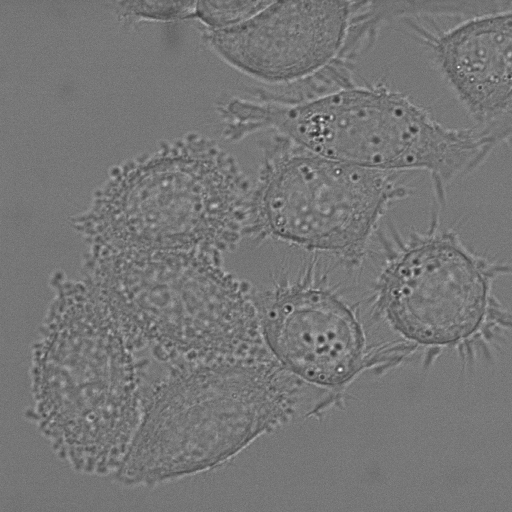}
    \caption{}
    \label{fig:dataset_ip_img}
\end{subfigure}%
\begin{subfigure}{0.3\textwidth}
    \centering
    \includegraphics[width=0.9\linewidth,keepaspectratio]{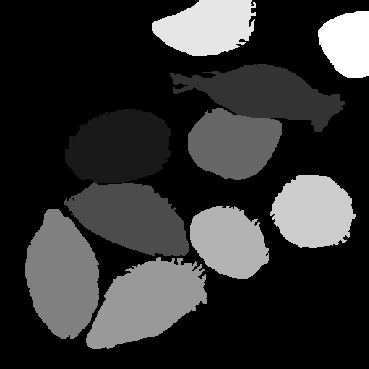}
    \caption{}
    \label{fig:dataset_label}
\end{subfigure}
\begin{subfigure}{0.3\textwidth}
    \centering
    \includegraphics[width=0.9\linewidth,keepaspectratio]{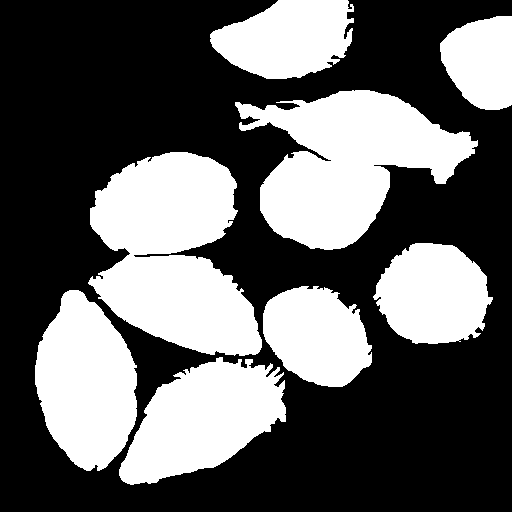}
    \caption{}
    \label{fig:dataset_label_2class}
\end{subfigure}

\caption{Example from the original ISBI cell dataset, (\ref{fig:dataset_ip_img}) Input image, (\ref{fig:dataset_label}) Ground truth with distinct classes for individual cells, and (\ref{fig:dataset_label_2class}) Ground truth with two classes}
\label{fig:dataset_img}
\end{figure}

The ISBI dataset consists of microscope images of biological cells, with tracking information for all images and segmentation information for some of the images. An example of input image and labeled segmented image from original dataset is shown in Figure \ref{fig:dataset_ip_img} and Figure \ref{fig:dataset_label} respectively. Originally, the individual cells in the ground truth images are coded to distinguish between different cells. For simplicity, we treat all cells as the same class, such that our system only needs to determine whether each pixel represents a cell or the background. The two-class representation of the ground truth can be seen in Figure \ref{fig:dataset_label_2class}.


The available ISBI cell dataset contains a total of 18 labeled segmentation images from two instances of tracking cell movements. While it would be tempting to use most of the 18 images to train the network and leaving out a few for testing, we instead use one of the two tracking instances for training (instance: 01) and the other one for testing (instance: 02). This allows for a less biased evaluation of the model using the test set, since all the cells images for training and testing are completely independent (rather than just temporally shifted images of the same cells). This results in 9 images for training and 9 images for testing. With very few full images available, data augmentation is necessary for the network to learn to generalize well. 

\begin{figure}
    \begin{center}
        \includegraphics[width=10cm,keepaspectratio]{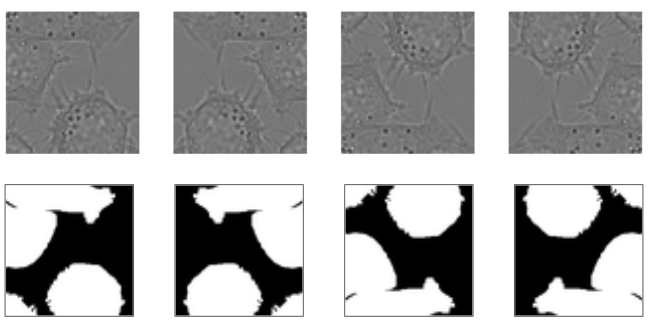}
    \end{center}
    \caption{Examples of train and test sample images with data augmentation.}
    \label{fig:data_augmentation}
\end{figure}

The data augmentation is performed online during training and evaluation. The model loads the dataset by parsing the input images related to labeled segmented images from the tracking instance. It further rotates ($0^\circ$, $90^\circ$, $180^\circ$, and $270^\circ$) the images and flips the image around the vertical axis, creating 45 images for each of the two tracking instances (01, and 02). These images are then randomly cropped to $192 \times 192$ from $512\times512$, and resized to $64 \times 64$. The ability to crop the image without affecting the content or statistics enables the augmentation to produce thousands of train and test images. 
 In order to train the network, we sample 20k images from generated dataset of 45 images from tracking instance 01 at every epoch with random cropping. The test set of 5k images is created using the generated dataset of 45 images from tracking instance 02. Some of the data augmented image samples along with respective ground truth from instance 01 are depicted in Figure \ref{fig:data_augmentation}.

\\
\subsection{Network Implementations}
\label{subsec:nw_implementation}
The simplified U-Net model is implemented as an ANN in two frameworks: TensorFlow (a standard deep network training library \cite{tensorflow2015}) and NengoDL (a library for deep learning applications that is part of the Nengo ecosystem \cite{rasmussen2018}). The NengoDL implementation can be run in spiking or non-spiking modes. It also allows direct porting to NengoLoihi, which provides a spiking network implementation that maps to the Loihi hardware.

The TensorFlow implementation is used to set a benchmark accuracy that is maintained while porting the network to NengoDL and NengoLoihi.  Different hyperparameters of the TensorFlow model (e.g., batch size, optimizer, activation function, dropout, kernel initializer, learning rate, etc.) are optimized using the Hyperopt library \cite{bergstra2015hyperopt} to achieve good accuracy. The optimized network (in TensorFlow) is used as a reference to compare the accuracy of the non-spiking implementation in NengoDL. With similar accuracy achieved via the TensorFlow and NengoDL DNN implementations, the SNN model in NengoDL is further optimized to realize the average firing rates of 50-200 Hz for individual layers.


To train the spiking network, we use an ANN-SNN conversion training method. The spiking network is trained using a rate-based nonlinearity that approximates the spiking nonlinearity. At inference (test) time, the rate nonlinearity is replaced with the target spiking nonlinearity. We choose the integrate-and-fire neuron model (with zero refractory period) for our spiking implementation, since it is supported by Loihi and typically trains well in practice. Using the approach pioneered by \cite{cao2014}, we train the ANN using the rectified linear unit (ReLU) nonlinearity, since this captures the firing rate of an integrate-and-fire neuron given the same input.

We use a modification of the ReLU nonlinearity during the forward pass to account for the fact that Loihi has a hard voltage reset (when a neuron spikes, the voltage is reset to zero). The nonlinearity we use in the forward pass is given by
\begin{align}
    p(x) &= \Delta t \lceil \frac{1}{x \Delta t} \rceil \\
    f_\text{forward}(x) &= \begin{cases}
      \left( p(x) \right)^{-1} & \text{if } x > 0 \\
      0 & \text{otherwise}
    \end{cases}
    \label{eqn:forward}
\end{align}
where $p(x)$ is the period between spikes (a.k.a. the inter-spike interval) for a given input $x$, $\Delta t$ is our simulation timestep (typically 1 ms), and $\lceil \cdot \rceil$ is the ceiling function. This equation has the effect of increasing the period between spikes to the nearest integer number of timesteps, for a given input. As $\Delta t \to 0$, this equation approaches the ReLU activation function $f(x) = \max(x, 0)$.

For the backwards pass, we ignore this equation (which is discontinuous even for $x > 0$), and use the derivative of the following equation:
\begin{equation}
    f_\text{backward}(x) = \begin{cases}
      \left( \frac{\Delta t}{2} + \frac{1}{x} \right)^{-1} & \text{if } x > 0 \\
      0 & \text{otherwise}
    \end{cases}.
    \label{eqn:backward}
\end{equation}
The effect of the $\Delta t / 2$ term is to unbias the equation with respect to the forward nonlinearity (Equation~\ref{eqn:forward}), so that the backward nonlinearity corresponds to the forward nonlinearity smoothed over the input $x$. Aside from this term, the backward nonlinearity is identical to the standard ReLU activation function $f(x) = \max(x, 0)$.

An additional modification we make to the forward pass is to add noise to the activation function to mimic the variability in the spiking output. This approach is described in detail in \cite{hunsberger2018}. We use a novel noise model that accounts for spikes being filtered by a first-order lowpass (a.k.a. exponential) filter with time constant $\tau$, and then integrated by the post-synaptic neuron's integrative membrane. The transfer function for this combined filter is
\begin{align}
    H(s) = \frac{1}{s(\tau s + 1)} \text{ .}
    \label{eqn:transfer}
\end{align}
We convolve this filter with a regular spike train of period $p$,
and analyze the case where the length of the spike train approaches infinity (i.e. the steady-state case):
\begin{align}
    s(t) &= \sum_{i=0}^{N \to \infty} \left( 1 - e^{-(ip + t)/\tau} \right) \nonumber \\
         &= N - \frac{e^{-t / \tau}}{1 - e^{-p / \tau}}
\end{align}
where $0 \leq t \leq p$ represents a point in time between one spike and the next.
Since this filter has infinite DC gain, 
the convolution results in an infinite series.
We therefore subtract the "ideal" response of the neuron
(i.e. the response of a ReLU neuron),
which is given by $((N - 1) p + t) / p$,
resulting in:
\begin{align}
    s(t) &= 1 - \frac{e^{-t / \tau}}{1 - e^{-p / \tau}} - \frac{t}{p} \text{ .}
\end{align}
Since the transfer function (Equation~\ref{eqn:transfer})
has a pure integrator in it,
this series is sensitive to offsets that happen at the beginning of the spike train, even when looking at infinite trains.
To account for this, we compute and subtract the mean of the series:
\begin{align}
    \mathbb{E}\left[ s(t) \right] &= \frac{1}{p} \int_0^p s(t) dt \nonumber \\
     &= \frac{1}{2} - \frac{\tau}{p} \text{ .}
\end{align}
To generate our training noise, we sample a random point from this series;
that is, we take $t = u p(x)$, where $u$ is sampled from a uniform distribution between zero and one.
The resulting nonlinearity for the forward pass is:
\begin{align}
    t &= u p(x) \nonumber \\
    \eta &= \frac{1}{2} + \frac{\tau}{p(x)} - \frac{e^{-t / \tau}}{1 - e^{-p(x) / \tau}} - \frac{t}{p(x)} \nonumber \\
    f_\text{forward}(x) &= \begin{cases}
      (1 + \eta) \left( p(x) \right)^{-1}  & \text{if } x > 0 \\
      0 & \text{otherwise}
    \end{cases}
    \label{eqn:forwardnoise}
\end{align}
This noise is ignored in the backwards pass;
that is, Equation~\ref{eqn:backward} remains the same.

During inference, we replace this rate-based nonlinearity
with the spiking integrate-and-fire neuron nonlinearity,
implemented on Loihi.
We use an exponential synaptic filter with $\tau = 5$ ms between layers.

\\
\subsection{Firing Rate Regularization}
\label{sec:firing_rate_reg}
The firing rates of neurons in an SNN impact the power and performance of the network significantly. If a neuron is spiking slowly, it takes longer to propagate information to the next layer of the network. If a neuron is spiking rapidly, on the other hand, performance improves but we lose the advantages of temporal sparsity. This is particularly problematic on Loihi, where the number of synaptic updates is directly proportional to the number of spikes going into a layer, and synaptic updates have a significant cost for both processing time and power consumption. Furthermore, the fact that Loihi neurons reset to zero after a spike means that given a constant input, a neuron will always fire with a period that is a constant integer number of timesteps. This means that at firing rates closer to the maximum rate of the inverse of the timestep $\frac{1}{\Delta t}$, there is less resolution (for example, if $\Delta t = 1\text{ ms}$, then a neuron can either fire at 500 Hz or 1000 Hz, but not at any rate in between). Based on these criteria, we have found that maximal firing rates in the range of 50-200 Hz work well to balance accuracy with energy efficiency and throughput.

To achieve this desired range of firing rates, we add a regularization term to our loss function that penalizes firing rates that are too high or too low. The most straightforward way to implement such regularization is with a standard regularization metric (such as L1 or L2 loss) that minimizes the distance between all neuron firing rates and a fixed target firing rate. However, this type of regularization can be in opposition to network accuracy, since it forces neurons to have similar firing rates for different inputs instead of allowing the neurons to use different rates to represent and process different inputs.

Instead of using a fixed firing rate target for all neurons on all examples, we make two key modifications: we allow a range of firing rates, and we regularize a rank-based statistic computed across a neuron's firing rates on multiple examples.
We propose the following percentile-based loss function to regularize the (almost) maximum firing rate of each neuron across all examples in the batch to be between a minimum $F_\text{min}$ and a maximum $F_\text{max}$ value:
\begin{align}
  \mathcal{L}^j_{FR} = \frac{1}{L_j}\sum_{i=1}^{L_j} \left( \left[ F_\text{min} - R^j_{i,p} \right]^+ + \left[ R^j_{i,p}-F_\text{max} \right]^+ \right)^2 \label{eqn:l2_percentile_loss}
\end{align}
where $L_j$ represents number of neurons in layer $j$,  $R^j_{i,p}$ represents the $p^\text{th}$ percentile of firing rate across a batch of input for neuron $i$ in layer $j$, and $[\cdot]^+$ is the ramp function. This equation has the effect of penalizing $R^j_{i,p}$ with a squared penalty if it is less than $F_\text{min}$ or greater than $F_\text{max}$, but otherwise imposing no penalty if it is within the acceptable range. If $p = 100\%$, then the equation regularizes the actual maximum firing rate of the neuron across all examples in the batch. In practice, we use $p = 99\%$ to allow the regularization to be more robust to outliers in the firing rate. We choose $F_\text{min} = 50 \text{ Hz}$ and $F_\text{max} = 200 \text{ Hz}$.
\\
\subsection{Loihi Implementation}
\label{subsec:loihi_implementation}

The network is deployed using NengoLoihi on either a simulated or real Loihi hardware platform.\footnote{A basic convolutional neural network example for NengoLoihi can be found at \url{https://www.nengo.ai/nengo-loihi/examples/cifar10-convnet.html} to give a sense of the conversion process.}

Here, we detail the two most important steps for implementing the network on Loihi: quantizing the network parameters to improve accuracy on Loihi, and partitioning the network to improve performance (i.e., energy efficiency and throughput) on Loihi.
\\
\subsubsection{Quantization}
\label{sec:quantization}

The neuromorphic processing cores on Loihi do not have floating-point arithmetic units; they do all their processing with fixed-point/integer arithmetic. Furthermore, they have a very specific architecture designed for energy efficiency. When quantizing network parameters for Loihi, it is important to quantize not only to the correct bit width, but to do so in a way that accounts for the specific Loihi architecture.

The synaptic and neuron compartment updates on Loihi can be summarized by the following equations:
\begin{align}
    \bar w_{ij} &= m_{ij} 2^a\\
    q_i[t] &= \sum_j \bar w_{ij} s_j[t]\\
    u_i[t] &= \left( u_i[t-1] \left(2^{12} - \delta_u\right) \right) \gg 12 + q_i[t] \label{eqn:u}\\
    v_i[t] &= \left( v_i[t-1] \left(2^{12} - \delta_v\right) \right) \gg 12 + u_i[t] + \bar b_i\label{eqn:v}
\end{align}
where a given synaptic weight $\bar w_{ij}$ is composed of a mantissa $m_{ij}$ and common exponent $a$; $s_j[t]$ is the incoming spikes on axon $j$ at time $t$ (equal to one if there is a spike, and zero otherwise); $\bar b_i$ is a fixed bias current for neuron $i$; $\delta_u$ and $\delta_v$ are the decay factors for the input current $u_i$ and membrane voltage $v_i$, respectively; and $x \gg n$ is the right-shift operation, which is equivalent to $\lfloor x 2^{-n} \rfloor$. When a neuron's voltage surpasses the discrete firing threshold $\bar v_\text{th}$, then the neuron fires a spike and the voltage is reset to zero. Note that these equations leave out many additional features of the Loihi chip that are unused by our network, as well as implementation details that are handled by NengoLoihi and are unnecessary for describing our quantization approach.

Given these equations, our goal is to determine a discrete firing threshold $\bar v_\text{th}$, a weight scaling factor $c$, and a weight exponent $a$, such that our floating-point weights $w_{ij}$ can be mapped to discrete weights by $m_{ij} = w_{ij} c$. The values of these three parameters must fulfill the following criteria:
\begin{itemize}
    \item the discrete network must exhibit the same behaviour as the continuous network as much as possible (i.e., approximately the same neuron firing rates for any set of inputs at each layer);
    \item the current and voltage for any neuron should not exceed the maximums representable by the chip, $u_\text{max}$ and $v_\text{max}$;
    \item the firing voltage $\bar v_\text{th}$ for all neurons should otherwise be as high as possible, to utilize as many of the available bits as possible when computing voltage decay;
    \item the weight mantissas $m_{ij}$ should fall in the allowable range of [-255, 255], and utilize as much of this range as possible to maximize the precision of the represented weights;
    \item the discrete bias $\bar b_i$ should fall below the maximum representable bias $b_\text{max}$ for all neurons.
\end{itemize}
The values are connected by the following equations:
\begin{align}
    \bar v_\text{th} &= v_\text{th} \frac{2^a c}{y[n]} \label{eqn:vbarth}\\
    \bar m_{ij} &= w_{ij} c\\
    \bar b_i &= b_i \frac{2^a c}{\hat y[n]}
\end{align}
where $\hat y[n]$ is the (normalized) integral of the total decayed response of the input current $u[t]$ to an input $u[0]$ ($y[n]$ is defined explicitly below in Equation~\ref{eqn:yn}). The fact that $\bar v_\text{th}$ is proportional to $c$ and $2^a$ indicates if we scale up our weights, we need to scale up our firing threshold accordingly to achieve the same behaviour. The fact that it is inversely proportional to $y[n]$ indicates that if our current decay $\delta_u$ is smaller, then the current will take longer to decay and an input of a given size will have a proportionally larger effect. The remainder of this section will detail how we determine the values in Equation~\ref{eqn:vbarth}.

We first determine $\hat y[n]$, which is an estimate of the integral of the input current $u[t]$ (Equation~\ref{eqn:u}) given an initial input current $u[0]$.\footnote{This analysis can also apply to the decay equation for $v[t]$ (Equation~\ref{eqn:v}), since it uses an identical decay mechanism. However, this is only necessary if the neuron type has a voltage decay; our neuron type for this work has no decay ($\delta_v = 0$).} This integral will allow us to determine the total effect of an input of a given size on the current $u[t]$. We begin by computing the discrete decay constant
\begin{align}
    \delta_u &= \left(2^{12} - 1\right) \left(1 - e^{-\Delta t / \tau_s}\right)
\end{align}
where $\Delta t = 1\text{ ms}$ is the length of one simulation timestep, and $\tau_s = 5\text{ ms}$ is the synaptic time constant. For a given input $u[0]$, we can then compute the decayed input $u[t]$ and its normalized integral $y[t]$ as follows:
\begin{align}
    r &= 2^{-12} \left( 2^{12} - \delta_u \right) \\
    u[t] &= \lfloor r u[t-1] \rfloor \\
    y[t] &= u[0]^{-1} \sum_{l=0}^t u[l]
\end{align}
We normalize the integral $y[t]$ to be independent of the input magnitude $u[0]$ with the intention of applying these results across inputs of varying magnitudes. The difficulty with computing $y[t]$ is that $u[t]$ contains a floor function, with the exact result depending on the specific input $u[0]$. We therefore replace $u[t]$ with an approximation $\hat u[t]$, where we replace the floor function with an expected ``loss'' due to rounding, $q$:
\begin{align}
    \hat u[t] &= r \hat u[t-1] - q \nonumber\\
       &= r^t u[0] - \sum_{k=0}^{t-1} q r^k \\
    \hat y[t] &= u[0]^{-1} \sum_{l=0}^t \hat u[l]
\end{align}
Assuming that $u[t]$ at any given time $t$ is a uniformly distributed integer, and $r$ is uniformly distributed between zero and one, then the fractional part of $r \cdot u[t]$ is also uniformly distributed between zero and one, and the expected loss due to rounding $q = 0.5$. However, we found empirically that $q = 0.494$ works better (since it accounts for effects such as an exactly zero remainder when $u[t]$ is a power of two greater than or equal to 12, which has an infinitely small probability when we assume uniformity of $r$). We can now compute the number of timesteps $n$ at which $\hat u[n] = 0$ for a given input $u[0]$:
\begin{align}
    \hat u[n] = 0 &= r^n u[0] - \sum_{k=0}^{n-1} q r^k \nonumber\\
    0 &= r^n u[0] - q \left(1 - r^n\right) \left(1 - r\right)^{-1} \nonumber\\
    n &= \frac{\log\left( q^{-1} (1-r) u[0] \right)}{\log r} \label{eqn:n}
\end{align}
Finally, we compute $\hat y[n]$:
\begin{align}
    \hat y[n] &= u[0]^{-1} \sum_{t=0}^n \left( r^t u[0] - \sum_{k=0}^{t-1} q r^k \right) \nonumber\\
    &= \frac{1 - r^{n+1}}{1 - r} - 
    \frac{q \left( n + 1 - (1 - r^{n+1}) (1 - r)^{-1} \right)}{(1 - r) u[0]} \label{eqn:yn}
\end{align}
Thus, given an input $u[0]$, we can approximate the total integral of that input (as processed by the fixed-point decay equation for $u[t]$), as:
\begin{align}
    \sum_{t=0}^\infty u[t] \approx \hat y[n] u[0] \text{ .}
\end{align}

\begin{figure}
    \begin{center}
        \includegraphics[width=12cm]{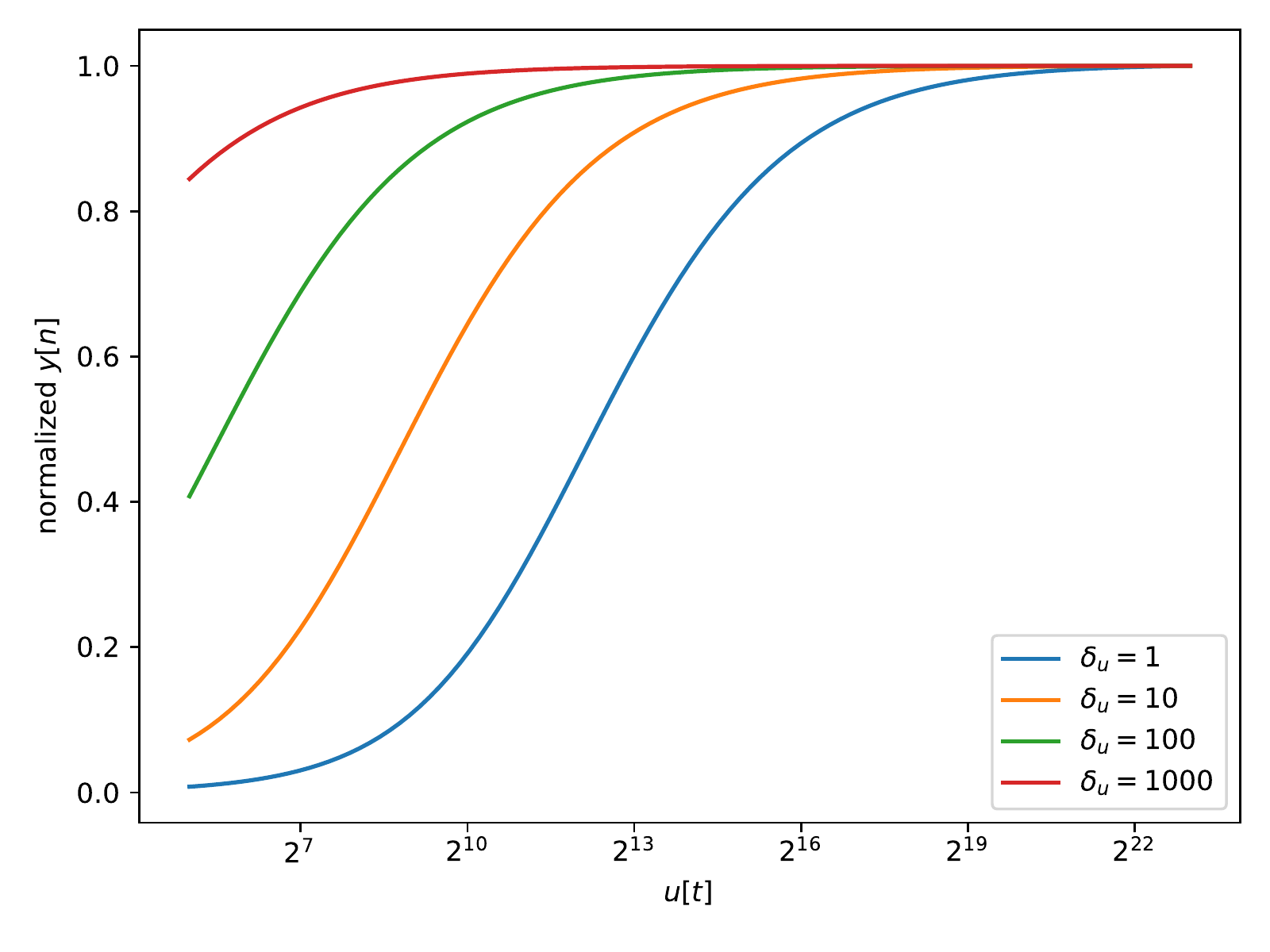}
    \end{center}
    \caption{Integrated input current $y[n]$ for various initial input currents $u[0]$. The x-axis shows the input $u[0]$, and the y-axis shows the integral $y[n]$ for that value of $u[0]$ normalized by the value of $y[n]$ for $u[0] = 2^{23}$. Individual traces show various decay values $\delta_u$ (for reference, when $\delta_t = 1\text{ ms}$ and $\tau_s = 5\text{ ms}$, as is standard in our network, then $\delta_u = 742$).}
    \label{fig:decay_magnitude}
\end{figure}

There are two caveats to this approximation. The first is that the computation of $y[n]$ is not independent of $u[0]$; we require $u[0]$ both when computing $n$ (Equation~\ref{eqn:n}) and directly in the equation for $y[n]$ (Equation~\ref{eqn:yn}). This means that we cannot compute a scaling factor $y[n]$ that will perfectly apply to all inputs. If we analyze these equations further, however, we can see that for all but the smallest decay values $\delta_u$ and the smallest input currents $u[0]$, the value of $y[n]$ is close to its asymptotic value that it approaches for very large $u[0]$. This can be seen in Figure~\ref{fig:decay_magnitude}. Therefore, by computing $y[n]$ for large $u[0]$, we should obtain a value that works well across the vast majority of inputs to the network. The other caveat is that we are assuming a linear superposition of inputs is possible, and that by computing $y[n]$ for one input, we can apply this to the case where there are many inputs at varying points in time. In the case of many inputs, we may be overestimating the amount of ``loss'' due to rounding, since the rounding only happens once per timestep for all inputs. The obvious alternative is to ignore the rounding and compute the integral as if there is no rounding, but this will over-estimate the integral particularly in the case where there is only one input, or inputs are sparsely distributed across time. In practice, we have found that it is better to assume that there is rounding ``loss'' than that there is none. 

We now select the weight scaling value $c$ such that the full range of floating-point weights $w_{ij}$ is mapped into the valid range of discrete weights [-255, 255]:
\begin{align}
    c &= \frac{255}{\max_{ij} \left| w_{ij} \right|} \text{ .}
\end{align}

The final value we need to determine our firing threshold $\bar v_\text{th}$ is the weight exponent $a$. We want to choose $a$ such that $\bar v_\text{th}$ is as large as possible (to best utilize the full bit width available on the chip for the neuron states $u[t]$ and $v[t]$, and thereby also minimize the influence of rounding during decay to those states), while also keeping $u[t]$ and $v[t]$ below their on-chip limits. To accomplish this, we start with $a = 6$ and progressively lower $a$ until $\bar v_\text{th} < v_\text{max}$ and $\bar b_i < b_\text{max}$.

\subsubsection{Partitioning}

We use the term ``partitioning'' to refer to selecting how the model is distributed across the available resources on the Loihi board. There are two main stages in our partitioning procedure:
\begin{enumerate}
    \item splitting groups of neurons that are too large to fit together on one neurocore;
    \item partitioning neurons on different cores to minimize inter-chip communication.
\end{enumerate}

For the splitting step, we take the neurons in each layer and split them into core-sized groups. We make the basic assumption that neurons that represent nearby parts of the image should be represented on the same core, all else being equal, since this allows a single input axon to update more neurons (with convolutional weights being local). Given that a layer is representing an $m \times n \times c$ image (where $c$ is the number of channels), we therefore try to find a $q \times r \times s$ region of the image to be represented on each core. This region must be small enough that it does not overuse the number of neurons, input or output axons, or weight memory available on the core. To find the ideal region size, we loop over all possible region sizes, and choose the one that results in the fewest cores being used to represent that layer\footnote{In the case of multiple configurations using the same minimal number of cores, we prefer the one that uses the fewest neurons per core (since this will best spread the computation across the cores).}, while satisfying the constraints.

To partition the cores across the two chips, we try to minimize the amount of communication between chips. Intra-chip communication on Loihi is significantly faster than inter-chip communication, and by minimizing the inter-chip communication we can significantly increase the throughput of our model. We create a graph of our model that represents each core as a node, and the number of axons connecting each pair of cores as an edge. We then use the METIS~\cite{metis} software library to find minimize the edge-cut of a 2-way partitioning of this graph, for which it employs multi-level methods and heuristics~\cite{karypis1998}.
\\
\subsection{Power Benchmarking Method}
\label{subsec:power_benchmark}
In order to provide the power consumption comparison between neuromorphic hardware (Loihi) and traditional hardware (CPU: Intel Xeon CPU E5-1650 and GPU: GeForce RTX 2080), we compute dynamic energy used during inference. This allows us to ignore the overhead (i.e., “idle power”) associated with running these devices from a workstation, and measure only the power associated with the computation, which is more representative of the power that would be used by an application-targeted system in the field. The CPU and GPU power benchmarking is performed on the rate-based ANN implementation in TensorFlow, since this implementation is more efficient on that hardware. The spike-based implementation is evaluated by measuring power consumption on Loihi (specifically on a Nahuku-32 board).\footnote{Although the network fits on two chips and therefore can be executed on a smaller Kapoho Bay board, a hardware bug on the older boards available to us does not allow the network to produce meaningful results. Thus we measure accuracy and power on a Nahuku-32 board instead. Newer Kapoho Bay boards do not have this issue.} The Nahuku-32 board has 32 Loihi chips, which results in much higher idle power than is necessary for this application (since all chips are idling, despite only two being used). Since we subtract out the idle power, our results are independent of the number of chips on the board. Having the additional idling chips may add some variability to the results, though we found in practice that the variability between subsequent trials was relatively low.

The idle power consumption of conventional hardware devices (CPU and GPU) is measured at the start of a run. Subsequently, test data is read, and the saved model is loaded. A small waiting period is added between model loading and power measurements, enabling the system to distance itself from the power consumption incurred by the network setup. The difference between this measurement and idle power is used to measure dynamic power consumption.  Given the weights of the network on Loihi are quantized as 8-bit integers, we have also demonstrated the power consumption effects on GPU with 8-bit integer quantized weights. NVIDIA's  \cite{tensorrt} library is used to realize a quantized network. The regular TensorFlow network with 32-bit floating point is converted to an INT8 calibration graph. A small representative dataset of 100 batches is used to calibrate the graph in order to compute the quantization parameters. This graph and the quantization parameters are stored in order to perform power benchmarking later on.

The power measurement procedure for Loihi is very similar, but slightly modified to better reflect the idle power on Loihi. Specifically, the network is built first, which then is followed by network execution power measurement and idle power measurement using NxSDK APIs. As before, the idle power of the device is subtracted from the measured power during the inference in order to compute the dynamic power consumption. Along with the power measurements, we also record the number of inferences, execution time (in seconds), and dynamic energy (in Joules) for the given batch size.

\section{Results}
\label{sec:results}
This section report the quantitative network segmentation accuracy, and a qualitative visual comparison between the prediction and the ground truth labeled images. It is followed by exploring the impact of inter-chip communication on speed and power consumption with optimized and unoptimized partitioning methods. We also represent the benefits of regularizing firing rate of Loihi supported neurons in terms of the accuracy and energy consumption of the network on Loihi. Subsequently, we demonstrate the comparison of throughput and energy consumption of ANN on CPU and GPU, with SNN on Loihi.
\\
\subsection{Network Performance on the Cell Dataset}
\label{subsec:nw_performance}
We train two networks: a UNet with ReLU non-linearity for getting baseline accuracy and power measurements on traditional hardware (CPU and GPU), and a UNet with Loihi Spiking non-linearity using NengoDL for benchmarking power and accuracy on neuromorphic hardware (Intel-Loihi board). We train the networks with 20,000 samples generated from the tracking instance 01 of the cell dataset at each of the 50 epochs. Independent of the training dataset, we generate another 5000 samples from tracking instance 02,  using the proposed data augmentation techniques (discussed in section \ref{subsec:data_augmentation}), for comparing performance accuracy of the networks. To compare the performance of the network, accuracy of the tensorflow trained network with ReLU non-linearity is computed in rate mode and the NengoDL trained network with Loihi Spiking non-linearity is evaluated in rate mode (ANN) as well as in spiking mode (SNN).

\begin{figure}[h!]
    \begin{center}
        \includegraphics[width=15cm,keepaspectratio,trim={2cm 10cm 0 0},clip]{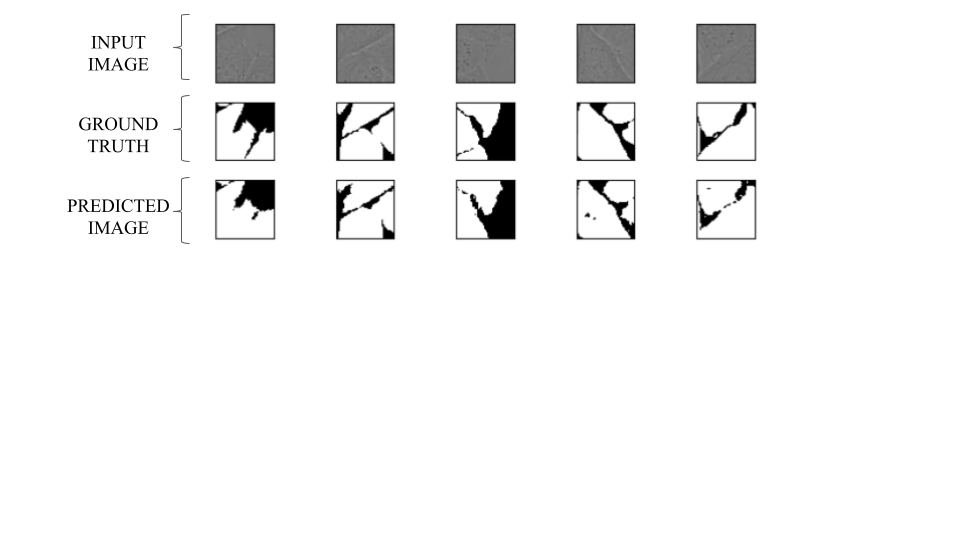}
    \end{center}
    \caption{Test sample prediction image and ground truth using ANN model inference in tensorflow.}
    \label{fig:ann_prediction}
\end{figure}

\begin{figure}[h!]
    \begin{center}
        \includegraphics[width=15cm,keepaspectratio,trim={2cm 7cm 0 0},clip]{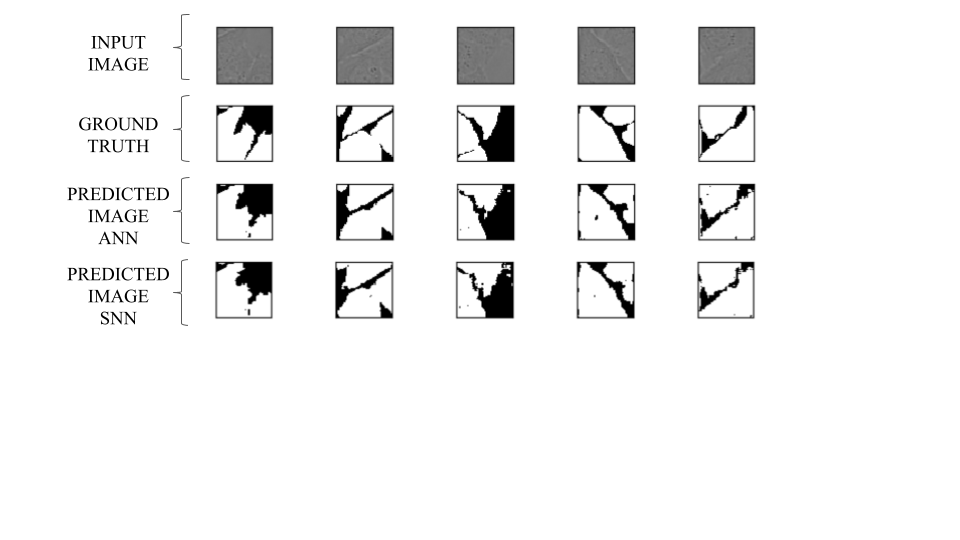}
    \end{center}
    \caption{Test sample prediction image and ground truth using Nengo-DL trained model inferred in rate mode (ANN) on GPU \& spiking mode (SNN) on Loihi.}
    \label{fig:loihi_prediction}
\end{figure}

We present the qualitative performance comparison of the networks by presenting the predictions of the networks and ground-truths side by side, for selected test images. The tensorflow trained network predictions appear to be fairly similar to ground-truths as shown in Figure \ref{fig:ann_prediction}. This similarity between predictions and ground-truths is also shared by the NengoDL trained network in both rate mode and spiking mode presented by  Figure \ref{fig:loihi_prediction}. Apart from predictions being similar to ground-truth, we also want the rate mode and spiking mode outputs to be as close as possible which can be observed in Figure \ref{fig:loihi_prediction}. Any differences between predictions of rate mode and spiking mode differences can be attributed to losses owing to quantization and spiking noise. 

We also present quantitative performance comparison of both networks using pixel accuracy as well as mean IoU. Pixel accuracy is estimated as the mean of percentage of correctly classified pixels, representing either background or cell, in the predicted output images. Mean IoU is calculated by dividing the intersection of predictions and ground truths pixels represented as cell class by the union of predictions and ground truths pixels represented as cell class. As demonstrated by Table \ref{tab:accuracy}, our baseline tensorflow trained network showcases high pixel accuracy of 94.98\% and mean IoU of 93.34\% on 5000 test samples of tracking instance 02. On the other hand, the NengoDL trained network in rate mode demonstrates slightly worse performance compared to baseline with pixel accuracy of 92.81\% and mean IoU of 90.34\%. We hypothesize that slight performance metrics reduction is a result of adding the firing rate regularizer when training the NengoDL network. Comparing the rate mode performance with spiking mode, the performance metrics in spiking mode are relatively unchanged, with a pixel accuracy and IoU that are both within 1\% of those of the ANN (92.13\% pixel accuracy and 89.58\% mean IoU).

\begin{table}[h!]
    \centering
    \begin{tabular}{|c|c|c|c|}
    \hline
        Metrics & \multicolumn{2}{c|}{ANN} & SNN\\\cline{2-3}
        & Tensorflow & Nengo-DL & \\
        \hline
        Pixel Accuracy & 94.98\% & 92.81\% & 92.13\%\\
        \hline
        mean IoU & 93.34\% & 90.34\% & 89.58\%\\
        \hline
    \end{tabular}
    \caption{Pixel accuracy and mean IoU measurement for ANN \& SNN network predictions}
    \label{tab:accuracy}
\end{table}

\subsection{Power Consumption}

\subsubsection{Partitioning}
Simplified UNet is the largest vision network implemented on Loihi. This multichip network involves intra-chip as well as inter-chip communication. The rate of inference depends on the speed of intra-chip and inter-chip communication. While Loihi is optimized for high speed intra-chip communication, we observe that the speed of inter-chip communication can significantly affect the throughout of our multichip network. In order to minimize the inter-chip communication, we have proposed a partitioning method using a serial graph partitioning library METIS.

\begin{figure}[h!]
    \begin{center}
        \includegraphics[width=16cm, keepaspectratio]{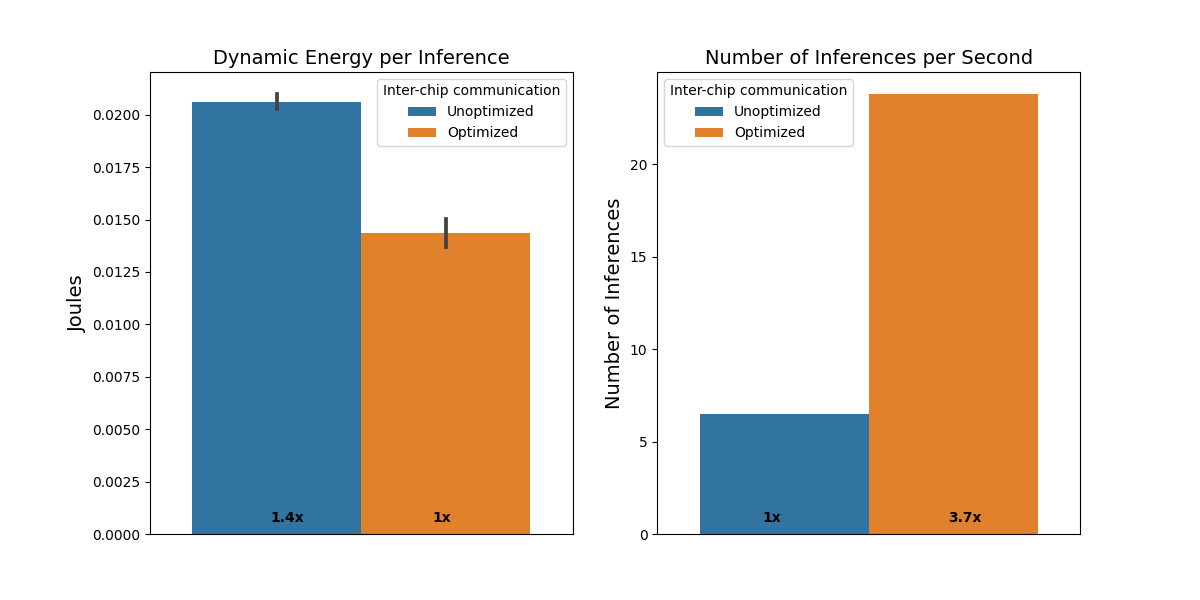}
    \end{center}
    \caption{Comparison of Dynamic energy per inference and throughput for optimized vs unoptimized inter-chip communication partitioning methods on Loihi}
    \label{fig:parition_power_compare}
\end{figure}

We investigated two networks, partitioned such that one is unoptimized and another is optimized to have minimum inter-chip communication. The unoptimized network has 8,130 inter-chip connections whereas optimized network has 3,739 inter-chip connections. The unoptimized network with larger number of inter-chip connections resulted in increased inter-chip communication and in turn higher execution time. This higher execution time of the unoptimized network resulted in lower throughput and higher energy consumption per image compared to the optimized network. Figure \ref{fig:parition_power_compare} demonstrates the comparison of dynamic energy per inference and throughput for both networks. We notice that the optimized network is consuming 1.4$\times$ less energy per inference and processing 3.7$\times$ more images per second compared to the unoptimized network.

\subsubsection{Firing Rate Regularization}
\label{subsec:results-firing-rate-reg}

Another significant factor for power consumption and network performance is the firing rate of the network. As discussed in Section~\ref{sec:firing_rate_reg}, we have found that maximum firing rates between 50 Hz and 200 Hz work well to balance accuracy with energy efficiency and throughput. 

One way to control the firing rate in a network is to modify the neuron amplitude, where the amplitude is a scaling factor on the output of the neuron. When implementing a spiking network, the amplitude can be folded into the efferent weights, such that the spikes themselves still have unit amplitude; therefore, changing the amplitude for all neurons has the effect of scaling the weights, and will result in different initial firing rates for the neurons. Without regularization, the network is not constrained to keep these firing rates, but rather they can freely change during training. In practice, we find that these initial firing rates do affect the final firing rates learned by the network, with smaller amplitudes qualitatively leading to higher firing rates, though the relationship between initial and final firing rates is difficult to quantify.

\begin{figure}[h!]
    \centering
    \includegraphics[width=16cm, keepaspectratio]{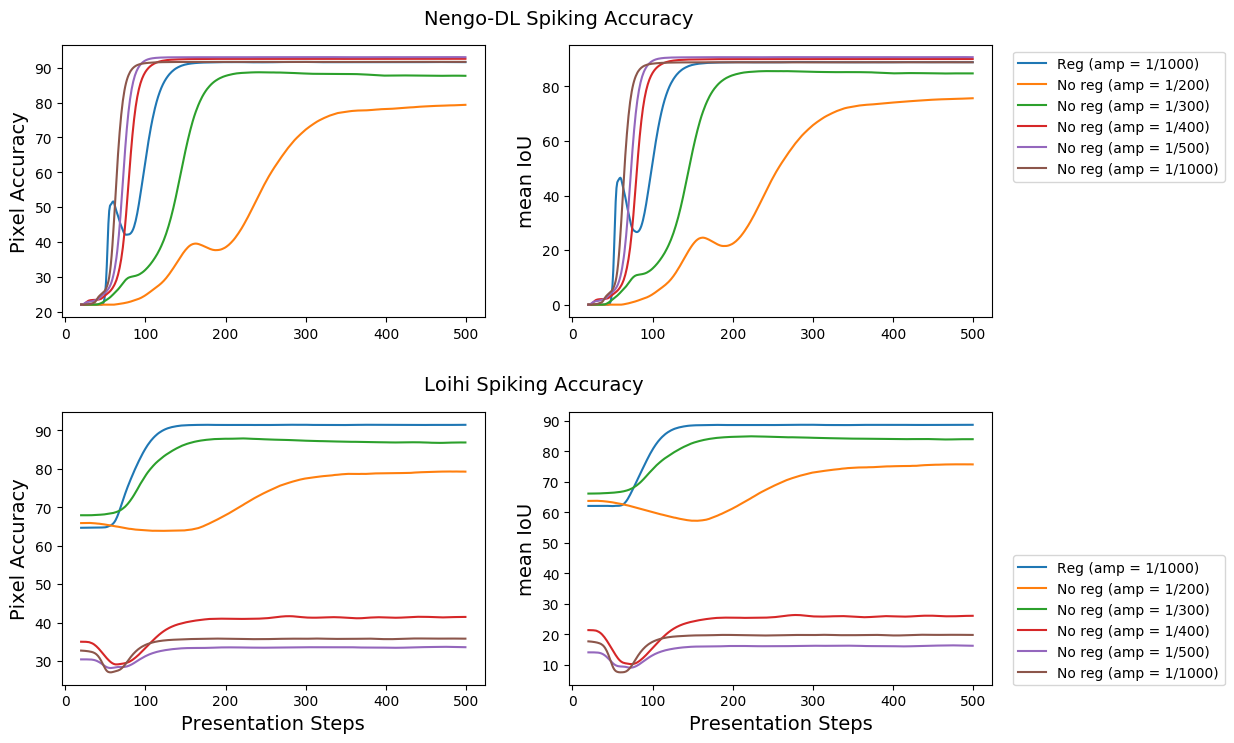}
    \caption{Comparison of accuracy over number of timesteps for firing rate regularized and unregularized networks.}
    \label{fig:frreg_acc_compare}
\end{figure}

To investigate the effects of firing rates on network performance, we trained networks without any firing rate regularization using five different neuron amplitudes: $\frac{1}{200}$, $\frac{1}{300}$, $\frac{1}{400}$, $\frac{1}{500}$, and $\frac{1}{1000}$. We measured the accuracy of these networks both in NengoDL and on Loihi. The NengoDL simulator uses spiking neurons and captures the hard-reset behaviour of the Loihi neurons (as described by Equation~\ref{eqn:forward}), but does not otherwise account for the quantization behaviour of Loihi (as described in Section~\ref{sec:quantization}). Figure \ref{fig:frreg_acc_compare} demonstrates the accuracy of the regularized and unregularized networks over time. As expected, networks with higher firing rates are able to achieve higher accuracy more quickly, due to the faster propagation of information through the network. Surprisingly, the accuracy of the unregularized networks with higher firing rates drops drastically when they are implemented on Loihi.

\begin{figure}[h!]
    \centering
    \includegraphics[width=10cm, keepaspectratio]{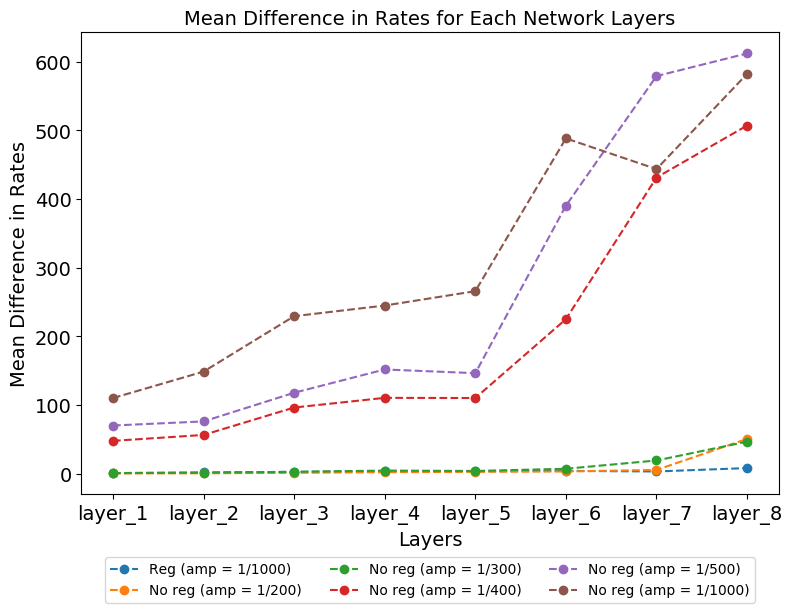}
    \caption{Comparison of layer-wise difference of mean firing rates between NengoDL and Loihi for networks with different firing rates. Rates are computed across 10 different stimuli. The unregularized networks with high firing rates show the largest differences. Our regularized network shows the smallest differences.}
    \label{fig:frreg_rate_diff}
\end{figure}

We believe these discrepancies in accuracy between NengoDL and Loihi for the higher-firing-rate unregularized networks results from the interplay between the Loihi neuron response curve and the quantization involved when implementing a network on Loihi. Due to the hard voltage reset used by Loihi neurons, the firing rate of a neuron can be highly discontinuous with respect to the input, specifically for higher firing rates (e.g., with $\Delta t = 1\text{ ms}$, an input current of 499 would result in a firing rate of 300, whereas an input current of 500 would result in a firing rate of 500). While this is accounted for by the NengoDL model, it does not account for the quantization involved with implementing the model on Loihi. Our hypothesis is that the variance caused by quantization has a much larger effect for networks with high firing rates, resulting in the higher error.  In support of this hypothesis, we found that there is a noticeable difference in the layer-wise mean firing rates between inference performed in NengoDL and inference performed on Loihi, specifically for the unregularized higher-firing-rate networks which showcased larger discrepancies in accuracy (Figure~\ref{fig:frreg_rate_diff}). We also note that the layer-wise mean firing rate difference between NengoDL and Loihi is lowest for our regularized network compared to others.

\begin{figure}[h!]
    \centering
    \includegraphics[width=16cm, keepaspectratio]{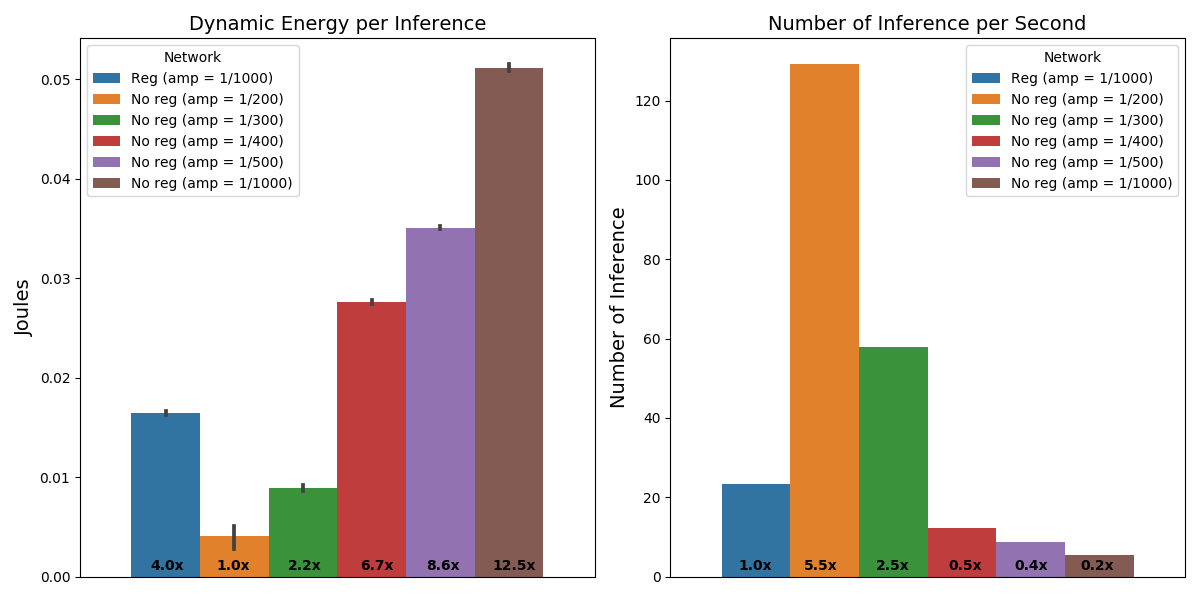}
    \caption{Comparison of dynamic energy per inference and throughput for firing rate regularized and unregularized networks for inference of 200 timesteps.}
    \label{fig:frreg_power_compare}
\end{figure}

We also compared the energy efficiency and throughput of the regularized and unregularized networks, as shown in Figure~\ref{fig:frreg_power_compare}. The energy consumption per inference correlates strongly with the mean firing rate of the network, and the throughput (in inferences per second) negatively correlates with the mean firing rate; this is expected, since processing more spikes takes more energy and more time. It is important to note that this assumes a fixed inference time of 200 timesteps, after which the pixel accuracy of the regularized network is $91.41\%$ whereas the lower-firing-rate unregularized networks have only achieved accuracies of $67.94\%$ and $87.83\%$ on Loihi (for the networks with amplitudes of $\frac{1}{200}$ and $\frac{1}{300}$, respectively).

\subsubsection{Spiking vs Non-Spiking Hardware} 
\label{subsec:power_consumption}

When comparing power consumption on different types of hardware, it is important to consider the batch size---that is, the number of examples that are run in parallel through the network. For many target applications, the input arrives “online,” one image at a time. For such applications, processing data in parallel is not possible, or has significant drawbacks in terms of latency (since the system must wait for a number of inputs to arrive to process them in parallel). Thus, our first comparison is for processing data online, i.e., a batch size of one. 

We investigated the power requirements, energy consumption and throughput for simplified UNet on traditional non-spiking hardware CPU and GPU, and spiking hardware Loihi for different batch sizes. In order to minimize the variance in power readings, we perform 15 trials for each experiment presented in this section \ref{subsec:power_consumption}. 

\begin{figure}
    \begin{center}
        \includegraphics[width=16cm, keepaspectratio]{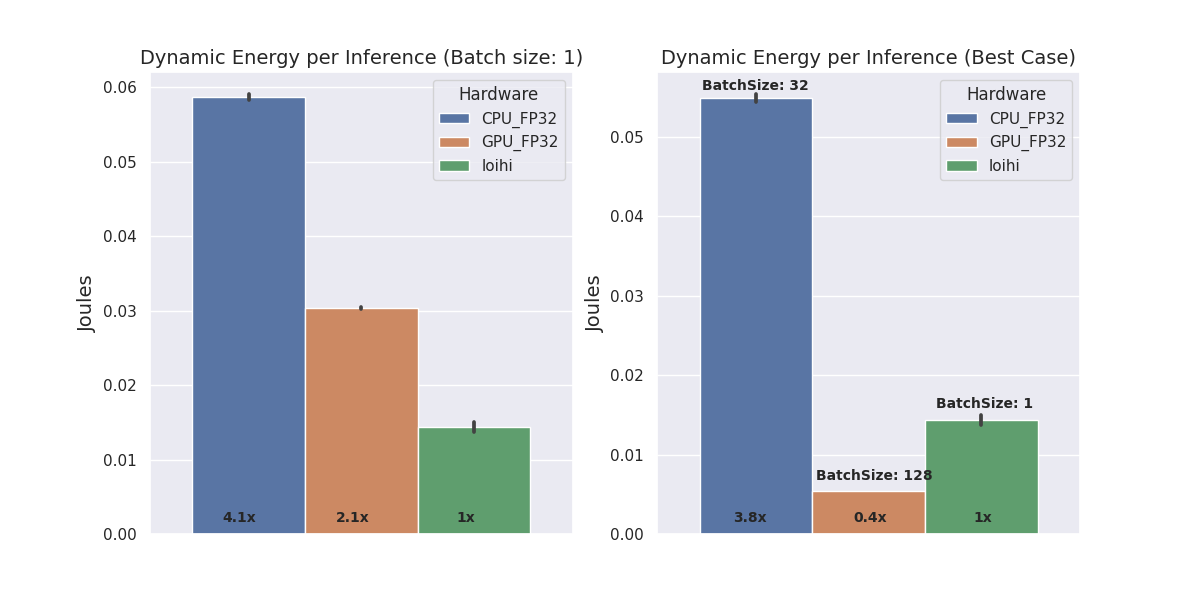}
    \end{center}
    \caption{Dynamic energy per inference comparison between CPU, GPU and Loihi for batch size of 1 and least energy per inference with optimized batch size}
    \label{fig:power_cpu_gpu_loihi}
\end{figure}

The energy consumption comparison is illustrated in Figure \ref{fig:power_cpu_gpu_loihi} (left). We see that Loihi uses 4.1x less energy than a CPU and 2.1x less energy than a GPU for a batch size of one. We also summarize the quantitative power performance comparison in Table \ref{tab:power_comparison} using inference per second and energy consumption per inference for CPU, GPU and Loihi. We observe relatively lower variance in different measurements over multiple trials which suggests the robustness of our experiments. A notable difference can be observed between the high running power of CPU and GPU compared to Loihi. A lower running cost and lesser energy consumption per inference makes Loihi suitable for many embedded applications. To some extent, this is unsurprising as Loihi was designed for online use, whereas GPUs and CPUs are being underutilized for small images so overhead costs may dominate. For this reason, we also examined larger batch sizes.

\begin{table}[ht!]
\def\arraystretch{1.5}
    \caption{ Mean power consumption and energy cost per inference across CPU, GPU and Loihi for batch size of 1. \\}
    \centering
    \begin{tabular}{|c|c|c|c|>{\centering\arraybackslash}m{3.1cm}|>{\centering\arraybackslash}m{2.5cm}|}
    \hline
    Hardware & Idle (W) & Running (W) & Dynamic (W) &   Inferences / Second   & Energy / Inference (J) \\
    \hline
CPU & 11.71 $\pm$ 0.05 & 37.01 $\pm$ 0.20 & 25.30 $\pm$ 0.19 & 431.44 $\pm$ 9.73 & 0.06 $\pm$ 0.00\\
    \hline
GPU & 3.21 $\pm$ 0.21 & 66.53 $\pm$ 0.78 & 63.32 $\pm$ 0.79 & 2086.55 $\pm$ 18.44 & 0.03 $\pm$ 0.00\\
    \hline
Loihi & 1.13 $\pm$ 0.03 & 1.47 $\pm$ 0.05 & 0.34 $\pm$ 0.03 & 23.79 $\pm$ 0.01 & 0.01 $\pm$ 0.00\\
    \hline
    \end{tabular}
    
    \label{tab:power_comparison}
\end{table}

On the GPU, the dynamic energy per inference decreases significantly as the batch size increases, since GPUs are designed for processing data in parallel. Even on the CPU, there can be some benefit to data parallelism, due to more efficient memory access and SIMD instructions. We investigated the effects of increasing batch size by performing multiple iterations of CPU and GPU inference for batch sizes of 2, 4, 8, 16, 32, 64, and 128. Notably, Loihi does not have any built-in support for data parallelism. While we could run multiple copies of the network in parallel on larger Loihi boards like Nahuku, there is no reason to expect a change in efficiency. To compare results across hardware platforms, the best performing batch size was chosen for both the CPU and GPU. Figure \ref{fig:power_cpu_gpu_loihi} (right) shows Loihi compared with the CPU using the batch size of 32, and the GPU using the batch size of 128.  Loihi performs better than the CPU, using over 3.8x less energy per inference for these larger batch sizes. Due to increased parallelism, the GPU consumes the 0.4x energy per inference compared Loihi for a large batch of 128. Recall that such large batch sizes introduce significant delay into the processing pipeline. For example, if the input is captured online i.e. one at a time, collecting 128 input samples will approximately incur 128x latency before it can be processed with GPU. This increased wait period may not be ideal for real-time online applications requiring to process the input sample as soon as received.

\begin{figure}
    \begin{center}
        \includegraphics[width=16cm, keepaspectratio]{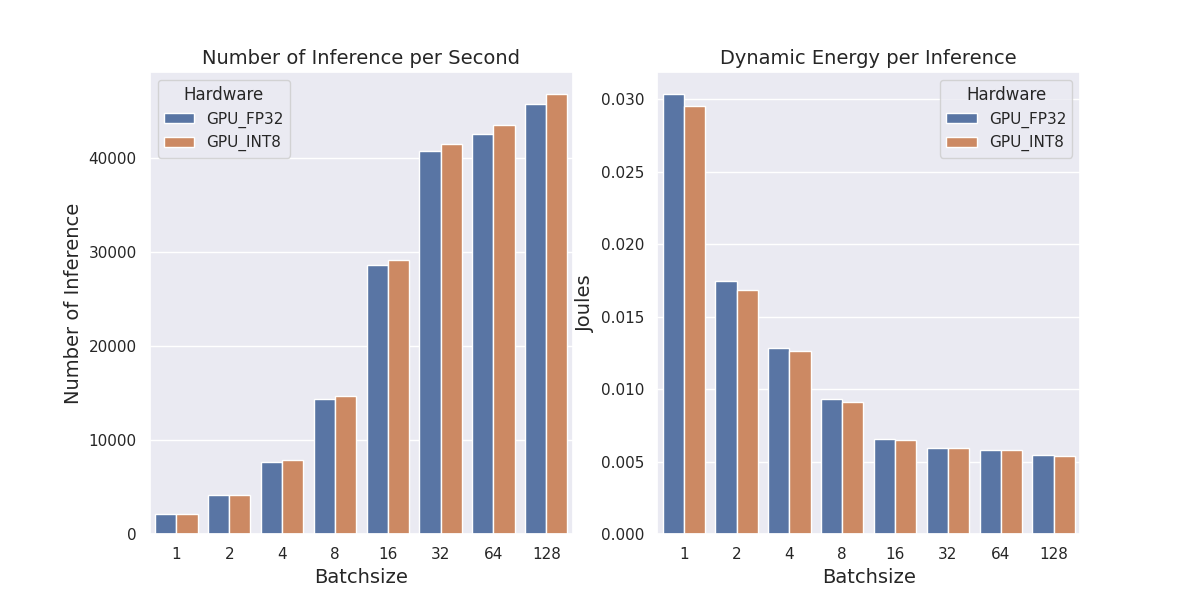}
    \end{center}
    \caption{Number of inference per second (left) and Dynamic energy per inference (right) comparison between regular FP32 and quantized INT8 network on GPU for different batch sizes.}
    \label{fig:power_gpu_quant}
\end{figure}

In order to test whether quantization of weights, required for Loihi, would also benefit the GPU, we have compared the GPU power consumption of two networks, one with 32 bit floating point weights (FP32) and another with 8 bit integer weights (INT8). We performed power measurements on the GPU for the FP32 and INT8 configurations with the batch sizes of 1, 2, 4, 8, 16, 32, 64 and 128 for 15 trials, and averaged the results in order to obtain energy consumption per inference. Figure \ref{fig:power_gpu_quant} shows the effect on the number of inferences per second (left) and dynamic energy per inference (right) for different batch sizes on the FP32 and INT8 networks. The number of inference per second as well as dynamic energy per inference are similar for the FP32 and INT8 networks for different batch sizes. We believe that the overhead computation introduced by quantization effectively overshadows the reduced computation benefits of a quantized network. Overall, these results suggest that the comparison between Loihi and INT8 network would be the same as the one with the FP32 network we presented earlier in this section.

\section{Discussion}
\label{sec:discussion}

The spiking U-Net presented in this paper has demonstrated the energy efficiency of neuromorphic computing for image segmentation.  We have directly compared two isomorphic networks and shown Loihi to run 2.1x more efficiently while preserving accuracy.  Our variant of the U-Net architecture is trained using NengoDL and TensorFlow to develop spiking and non-spiking versions.  The non-spiking network achieved a pixel accuracy of 94.98\% and 92.81\% on the test dataset with networks trained in TensorFlow and NengoDL, respectively (where the NengoDL network uses the Loihi neuron response curve and firing rate regularization). The spiking network achieves a similar pixel accuracy of 92.13\% on the same test dataset, running on the Loihi neuromorphic hardware with greater energy efficiency. The Loihi network uses the same number of neurons, the same network structure and connectivity pattern, and the same parameters as the ANN, with the exception that the parameters are quantized to be suitable for Loihi. Consequently, the comparison is appropriate for drawing conclusions about how the same network can run on different platforms.

While there may be concern that the GPU model is using a 32 bit representation and the Loihi is largely using 8 bits, we make two critical observations. First, the networks have been controlled for accuracy, so the comparison is of functionally similar networks. Second, and more importantly, we have extensively tested the quantized U-Net architecture and shown that such quantization does not significantly change power usage.

Critically, these two functionally and architecturally similar networks do not use the same amount of power. The power benchmarking demonstrates over 2x less energy consumption by the Loihi neuromorphic hardware compared to conventional hardware, when processing one image at a time, as is required of online inference.  This sort of processing is important for data streaming and real-time applications as it will introduce minimal latency into the response of the system.  Such functionality is important for real-time video processing, audio processing, rapid forecasting, control and many other applications.

While the original U-Net architecture is simplified to fit on two Loihi chips, there is nothing in the hardware, methods or tools that limits network size intrinsically.  In addition, even this simplified version of the U-Net remains one of the larger functional neuromorphic examples, and the largest vision network run on event-based neuromorphic hardware of which we are aware.

Our results demonstrate the importance of firing rate regularization and accounting for inter-chip communication when mapping an ANN to Loihi. Our network with regularized firing rates was able to achieve better accuracy on Loihi than any of the unregularized networks. Furthermore, without regularization the only way to determine the final network firing rates for a given network initialization is to train the network, which makes it quite tedious to target a particular firing rate regime without regularization. Regularization allows one to target a firing rate regime that is known to work well on Loihi, independently of the network initialization. We also found that partitioning the network in a way that reduces inter-chip communication not only lowered power consumption, but had a very dramatic effect on network throughput, increasing it by a factor of over 3$\times$.

We found that both power usage and throughput are highly correlated with the average firing rates of neurons in the network. This indicates that for our networks, much of the time and energy in the network is spent doing synaptic updates which are directly proportional to the number of spikes, rather than neuron updates which must be done every timestep regardless of the number of spikes. While networks with lower firing rates take longer to propagate information through the network, and thus more timesteps to achieve the desired accuracy, each timestep takes both less energy and less time to compute. There is a balancing act between more lower-power and faster timesteps with low-firing-rate networks, and fewer higher-power and slower timesteps with high-firing-rate networks. The only method we currently have to determine which is better for overall power and throughput is to empirically try different configurations. We do know that if firing rates are reduced too much, then neuron updates will begin to dominate energy usage and compute time, and there would be no advantage to reducing firing rates further. If firing rates are increased too much, then the hard-reset behaviour of Loihi neurons makes their response curves highly discontinuous, which results in a severe drop in accuracy when transferring these networks to Loihi (as shown in Section~\ref{subsec:results-firing-rate-reg}).

One important area for future research is to apply these methods to example applications that are time varying.  Past work using Loihi has shown particularly good results on time-varying speech data \cite{blouw2018keyword}. However, such data is low-dimensional and typically requires smaller networks to run.  Consequently, video processing would be a natural next step, and one for which latency is often a significant issue with currently available hardware. 

\section{Conclusion}

We have demonstrated improved energy efficiency on the U-Net architecture using the Loihi neuromorphic chip, while retaining accuracy. We believe this is one of the largest published examples of a functional application on neuromorphic hardware and the largest published example on Loihi specifically, demonstrating the presumed energy efficiency of neuromorphic hardware while maintaining performance.  While this application remains modest, we note that the tools, techniques and hardware employed can scale significantly beyond this initial application.  In this way, while the present work demonstrates that neuromorphic hardware can leverage biological inspiration to realize some of the energy efficiency observed in nature, much remains to be done.  In particular, the application to static image segmentation is not an ideal one to demonstrate the strengths of neuromorphic approaches.  Spike-based implementations are likely more efficient in applications with time-varying inputs, rather than static ones.  Nevertheless, this paper provides evidence of the usefulness of current tools and platforms for moving the field towards larger, more efficient, and highly functional applications of neuromorphic technology.

\bibliographystyle{unsrt}  
\bibliography{reference}  

\end{document}